\documentclass[journal]{IEEEtai}

\usepackage[colorlinks,urlcolor=blue,linkcolor=blue,citecolor=blue]{hyperref}

\usepackage{color,array}
\usepackage{float}

\usepackage{graphicx}
\usepackage{bibentry}
\usepackage[numbers]{natbib}  
\usepackage{hyperref}
\usepackage{caption} 
\usepackage{subcaption}
\usepackage{booktabs}
\usepackage{amsmath}

\begin{document}


\title{Label-Efficient School Detection from Aerial Imagery via Weakly Supervised Pretraining and Fine-Tuning}

\author{Zakarya Elmimouni, Fares Fourati, Mohamed-Slim Alouini
\thanks{The authors are with the \textbf{Computer, Electrical and Mathematical Sciences and Engineering (CEMSE) Division, King Abdullah University of Science and Technology (KAUST)}, Thuwal 23955-6900, Saudi Arabia (Email:\{zakarya.elmimouni, fares.fourati, slim.alouini\}@kaust.edu.sa) \textbf{Corresponding author:} Zakarya Elmimouni}
}

\maketitle

\begin{abstract}

Accurate school detection is essential for supporting education initiatives, including infrastructure planning and expanding internet connectivity to underserved areas. However, many regions around the world face challenges due to outdated, incomplete, or unavailable official records. Manual mapping efforts, while valuable, are labor-intensive and lack scalability across large geographic areas. To address this, we propose a weakly supervised framework for school detection from aerial imagery that minimizes the need for human annotations while supporting global mapping efforts. Our method is specifically designed for low‑data regimes, where manual annotations are extremely scarce. We introduce an automatic labeling pipeline that leverages sparse location points and semantic segmentation to generate infrastructure masks from which we generate bounding boxes. Using these automatically labeled images, we train our detectors on a first training stage to learn a representation of what schools look like, then using a small set of manually labeled images, we fine-tune the previously trained models on this clean dataset. This two stage training pipeline enables large-scale and strong detection in low-data setting of school infrastructure with minimal supervision. Our results demonstrate strong object detection performance, particularly in the low-data regime, where the models achieve promising results using only 50 manually labeled images, significantly reducing the need for costly annotations.
This framework supports education and connectivity initiatives worldwide by providing an efficient, extensible approach to mapping schools from space. All models, training code, and auto-labeled data will be publicly released to foster future research and real-world impact.

\end{abstract}

\begin{IEEEImpStatement}

School infrastructure mapping is a key component of global education and connectivity initiatives. However, many regions still rely on incomplete or outdated records, while manual annotation of aerial imagery remains costly and difficult to scale. This work introduces a weakly supervised detection framework that overcomes these limitations by combining automatic labeling from sparse geolocation data with a two-stage training strategy. The proposed approach significantly reduces the need for human annotations while maintaining strong object detection performance, enabling scalable deployment across large geographic areas. By improving the efficiency of representation learning under limited supervision, this framework can support large-scale mapping efforts such as the Giga Initiative. It is applicable to infrastructure monitoring, urban planning, and public policy, and can facilitate better resource allocation in underserved regions, contributing to more inclusive access to education and digital services worldwide.
\end{IEEEImpStatement}

\begin{IEEEkeywords}

School Detection, Automatic Labeling, Weakly Supervised Learning.
\end{IEEEkeywords}

\begin{figure*}[t]
    \centering
    \includegraphics[width=0.9\linewidth]{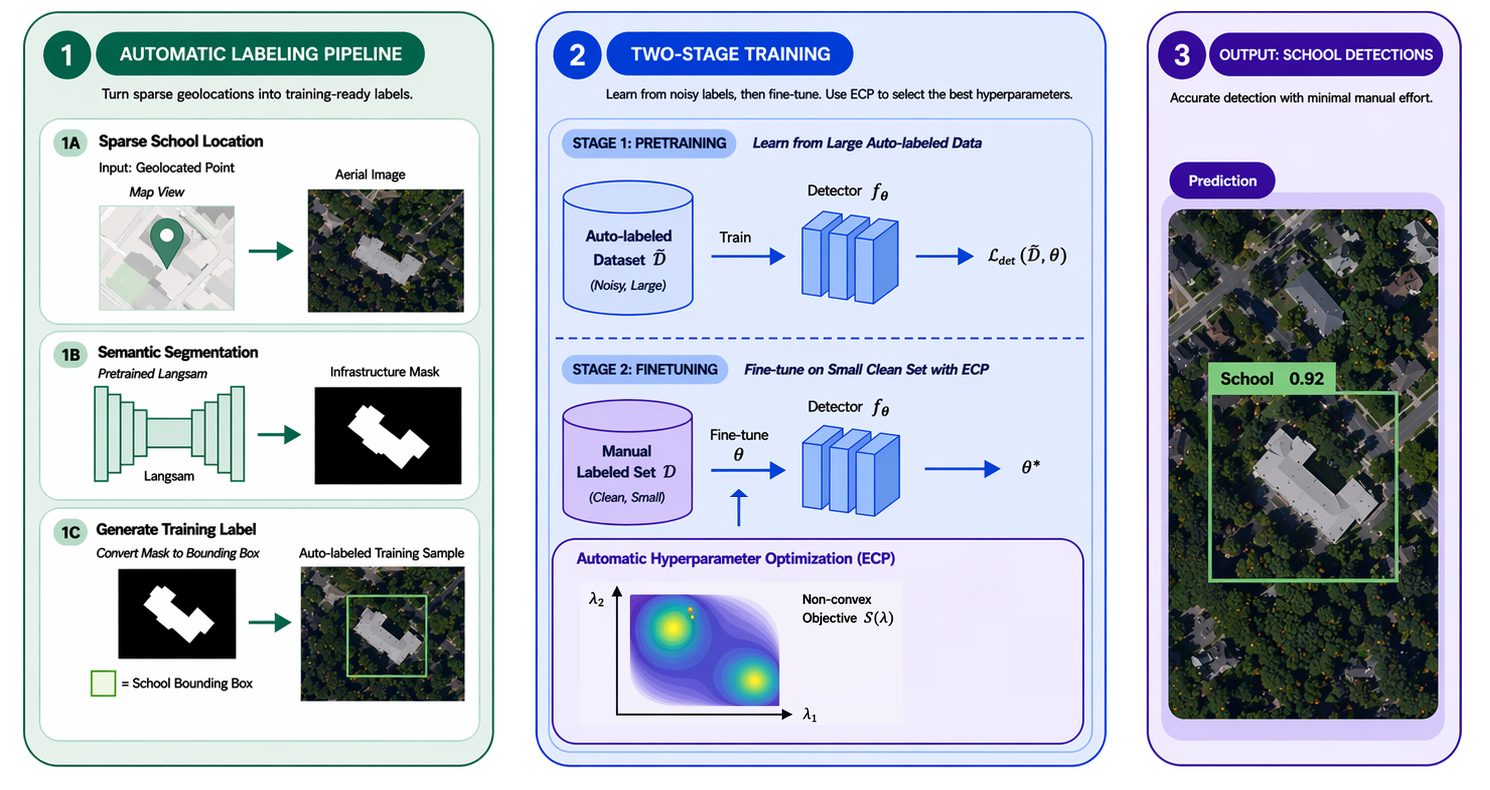} 
    \caption{\small Overview of the proposed weakly supervised pipeline for school detection from aerial imagery. Sparse geolocated points are first mapped to image regions, cleaned, and converted into bounding-box labels via semantic segmentation using LangSAM \citep{langsam}, yielding a large but noisy auto-labeled dataset. A detector (e.g., YOLO \cite{yolo12_ultralytics}) is then pretrained on this data to learn robust representations, and subsequently fine-tuned on a small, clean manually labeled set to improve localization accuracy. Hyperparameters are optimized with Every Call is Precious (ECP) \citep{fourati2025every}, which explores the configuration space under a non-convex validation objective.}

    \label{fig:diagram}
\end{figure*}

\newpage
\section{Introduction}
\IEEEPARstart{T}{he} 
 global learning crisis faces an urgent inflection point: nearly 3.7 billion people remain offline \citep{UN2021}, with children in many countries disproportionately excluded from digital opportunity as highlighted by the United Nations Children's Fund \citep{UNICEF2020}. For instance, in Latin America and the Caribbean, 49\% of children aged 3 to 17 remain unconnected \citep{UNICEFITU2020}. These stark regional disparities underscore the urgency of bridging the digital divide to guarantee inclusive and equitable education for all.

In response, UNICEF and the International Telecommunication Union (ITU) launched the Giga Initiative, aiming to connect every school to the internet by 2030 \citep{giga2025}. However, Giga faces a foundational challenge: in many regions, schools lack digital footprints altogether. For instance, in Kenya, only 7,000 of approximately 33,000 schools have verified geographic coordinates \citep{tingzon2025}, leaving the majority unaccounted for in infrastructure and connectivity planning.

Accurate mapping of school infrastructure from aerial and satellite imagery is a key enabler for large-scale education and connectivity initiatives. While several global efforts aim to improve the availability of geospatial data \citep{giga_data}, existing school location records are often incomplete, inconsistent, or difficult to maintain at scale. As a result, automated approaches based on computer vision have emerged as a promising solution to detect and localize schools directly from aerial and satellite imagery.

Existing computer vision methods for mapping schools from aerial and satellite imagery often fall short of real-world applicability, either due to insufficient localization precision or to the high cost of manual annotation. Tile-based classification approaches \citep{doerksen2024, maduako2022, yi2019} can identify regions that are likely to contain schools but do not provide the precise localization needed for infrastructure deployment. Object detection methods offer more accurate localization but typically rely on costly manual annotations, limiting their scalability across diverse geographic contexts \citep{fu2021}.


In this work, we address these challenges by proposing a weakly supervised detection framework that significantly reduces the need for manual annotations while preserving strong detection performance. The proposed label-efficient framework follows a two-stage training strategy that combines noisy and clean data in a principled manner. First, we automatically generate bounding boxes for the schools using sparse location data and semantic segmentation. Object detectors are then pretrained on this large automatically labeled dataset. In the second stage, the pretrained detectors are fine-tuned on a small, high-quality manually annotated golden dataset to refine localization accuracy and improve detection performance. As we show in the following sections, this approach achieves strong detection performance precisely where manual labels are limited, making it a scalable solution suitable for high impact initiatives such as the Giga project, where scalability is a critical requirement. For scientific completeness only, we also evaluate our framework with larger manually labeled subsets (up to 443 images); these experiments are not the focus of our work.

We evaluate our framework on a manually annotated dataset of school and non-school images and conduct a systematic comparison of multiple object detection architectures under different data regimes. Our contributions are:

\begin{itemize}
    \item \textbf{Label-Efficient School Detection from Aerial Imagery:} We present a framework for detecting school infrastructure in high-resolution aerial imagery that achieves strong performance with as few as 50-100 manually annotated images, significantly reducing annotation requirements.
    
    \item \textbf{Automatic Label Generation from School Location Data:} We propose a scalable method to annotate the images automatically by converting school geolocation points into bounding box annotations, enabling large-scale weak  annotations without any manual labeling.
    
    \item \textbf{Two-Stage Weakly Supervised Training Strategy:} We demonstrate that pretraining on automatically generated labels followed by fine-tuning on a small, high-quality manually annotated dataset consistently improves detection performance compared to training on limited labeled data alone.
    
    \item \textbf{Evaluation Across Models and Data Regimes:} We provide a comprehensive evaluation of the proposed framework across multiple object detection architectures and varying data regimes. Our primary focus is on low‑data scenarios (50–100 labels) relevant to large‑scale school mapping; larger regimes (300–443 labels) are included only for scientific reference.
\end{itemize}

Together, these contributions provide a pathway toward globally scalable school detection systems with minimal human annotation effort, supporting equitable digital infrastructure planning in alignment with initiatives such as GIGA. To promote reproducibility, facilitate future research, and maximize real-world impact, we make all model weights, training code, and generated annotations publicly available at \href{https://github.com/zakarya-elmimouni/School_Detection}{https://github.com/zakarya-elmimouni/School\_Detection}. 

\begin{figure*}[t]
    \centering
    \subcaptionbox{Example1}[0.21\linewidth]{
        \includegraphics[width=\linewidth]{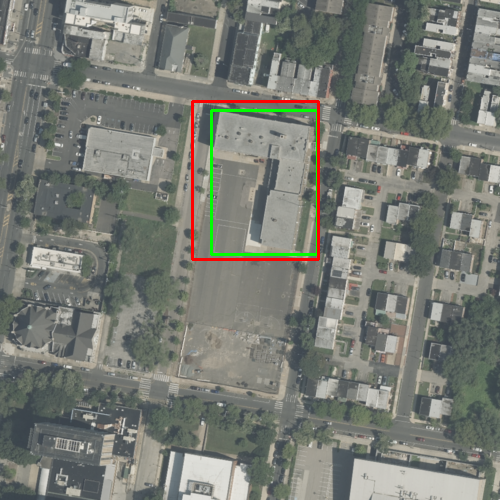}
    }
    \hfill
    \subcaptionbox{Example2}[0.21\linewidth]{
        \includegraphics[width=\linewidth]{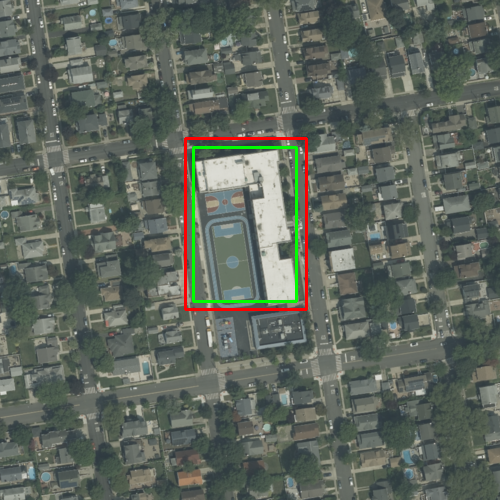}
    }
    \hfill
    \subcaptionbox{Example3}[0.21\linewidth]{
        \includegraphics[width=\linewidth]{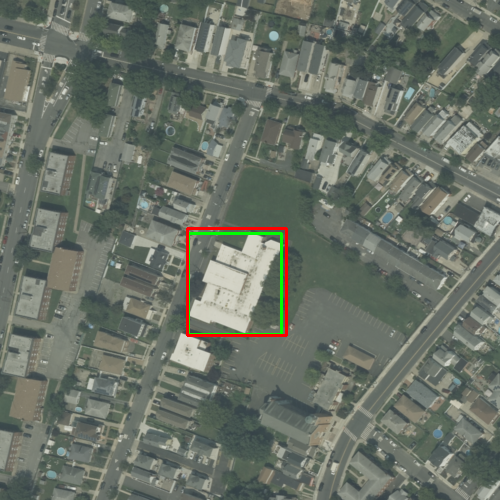}
    }
    \hfill
    \subcaptionbox{Example4}[0.21\linewidth]{
        \includegraphics[width=\linewidth]{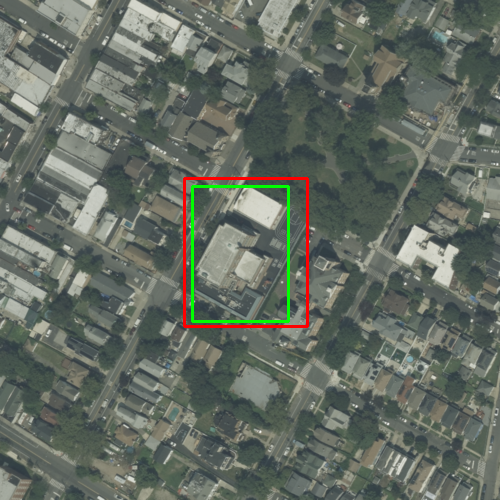}
    }

    \caption{Qualitative predictions of a detector across four examples. Ground truth is shown in green; predictions in red.}
    \label{fig:global_model_examples}
\end{figure*}

\section{Related Work}

Object detection is a core task in computer vision, involving both the identification and localization of specific object categories within complex visual scenes \citep{PAGIRE2025110075}. It has been widely applied across domains such as agriculture \citep{badgujar2024agricultural}, medical applications \citep{ragab2024comprehensive}, and remote sensing \citep{zhu2024scnet, xiao2025yolors, Sirko2021Continental}. Recent advances in detection architectures, including several YOLO-based models \citep{yolov10, yolo11_ultralytics, yolo12_ultralytics} and RCNN based models \citep{rcnn,fastrcnn,fasterrcnn} have significantly improved both accuracy and inference speed, enabling large-scale deployment in real-world settings.

School mapping in aerial and  satellite imagery has largely been approached as a tile-level classification task rather than as object detection. For example, \citep{Yazdani2018} proposed an unsupervised method to identify rural schools in Liberia by clustering ResNet-derived feature vectors using k-means \citep{likas2003global}. While this method effectively narrows the search space, it lacks precise localization capabilities, is highly sensitive to the choice of feature representation and clustering parameters, and depends on manual inspection, limiting both its scalability and level of automation.

Other studies similarly frame school mapping as a binary classification problem over image tiles \citep{yi2019, maduako2022, doerksen2024}. For instance, \citep{yi2019} and \citep{maduako2022} demonstrated the utility of Convolutional Neural Networks (CNNs) for large-scale school classification, but required extensive manual filtering of training data. While this improves model accuracy, it imposes significant labor costs and limits applicability to diverse regions. A key limitation across these classification-based approaches is the lack of precise localization, which is critical for infrastructure planning. In contrast, our method formulates school detection as an object detection task, producing accurate bounding boxes and enabling scalable deployment across geographies.

Beyond tile classification, \citep{tingzon2025} introduced a method for approximating school locations using Class Activation Maps (CAMs) derived from classification models. By selecting the most activated pixel in the CAM output, they estimate geographic coordinates for each detected school. While promising, this approach is inherently limited by the interpretability and reliability of activation maps. Since deep models are often opaque \citep{Amin2023}, there is no guarantee that the highlighted regions consistently correspond to actual school infrastructure.

A fully supervised object detection approach to school mapping was explored in \citep{fu2021}, where the authors trained object detectors using manually annotated bounding boxes across eight urban cities in China. While their model achieved good performance in those settings, the method relies entirely on costly human labeling, making it difficult to scale to other countries or rural contexts with limited annotation resources.




\section{Dataset Construction}



\subsection{School and Non School Coordinates Collection}

In this work, we focus on the United States as the country of interest. Although our approach does not inherently require a large number of training images, we selected the U.S. because of the availability of data in this country with sufficient scale, diversity, appropriate licensing, and good resolution, enabling a robust validation of our method.

Comparable high-resolution imagery datasets are often not publicly available for many other regions and typically require commercial access through data providers. For instance, some datasets, such as those provided through Google initiatives \citep{static_maps}, offer imagery for low- and middle-income countries (LMICs), but their licensing restrictions often prevent their use for training or redistribution, limiting their suitability for our study.


Although our experiments are conducted on the U.S. dataset, the proposed methodology is designed to be scalable and transferable to other geographic contexts. Therefore, the experiments presented in this work should be interpreted as a case study demonstrating the effectiveness of the proposed approach under realistic data availability constraints.

The GIGA dataset provides geolocated point indicating the locations (longitude, latitude) of schools. However, it does not include detailed object detection annotations such as bounding boxes or segmentation masks, which motivates the development of automated methods for generating such annotations at scale.


Positive samples (schools) positions were sourced from the GIGA initiative \citep{giga_data}. For negative samples (non school) positions, we adopted a strategy similar to \citep{doerksen2024} and \citep{tingzon2025}, using OpenStreetMap (OSM) and the Overpass API to collect geolocated features corresponding to non-school infrastructure, including hospitals, supermarkets, parking lots, and other public or commercial buildings. In contrast to \citep{maduako2022}, we did not sample from unpopulated or rural areas. Instead, negative examples were drawn exclusively from built-up regions, encouraging the model to learn fine-grained, discriminative features that differentiate schools from visually similar urban structures.

\subsection{School and non School Image Collection}
After retrieving the coordinates, we cleaned the dataset by removing duplicates and points that were too close to each other. Specifically, we enforced a minimum distance of 300 meters between any two points of interest, whether schools or non-schools, to remove overlap and ensure spatial diversity. For each remaining coordinate, we downloaded a 500×500 pixel  high-resolution aerial imagery from the National Agriculture Imagery Program (NAIP), which offers imagery at approximately 0.6-meter spatial resolution \citep{naip_dataset}.

\subsection{Golden Dataset: Manually Labeled Subset}

Among the images downloaded using school coordinates from the GIGA initiative, we manually constructed a high-quality \textit{golden} dataset to serve as \textit{the main source of ground truth for evaluation}. From the collected imagery, we selected a subset of images that clearly depicted school infrastructure and manually annotated precise bounding boxes around the visible school buildings. To ensure annotation reliability, only visually unambiguous cases were retained, while images with uncertain or ambiguous structures were excluded from the annotated set.

To help the models learn discriminative visual features and distinguish schools from other structures, we also included manually selected non-school images. These negative samples were drawn from previously collected non-school locations, with most examples originating from populated urban areas and a smaller portion sampled from visually diverse regions such as forests.

The resulting golden dataset contains manually annotated images and constitutes the only source of fully reliable ground-truth annotations used in this study. For evaluation purposes, the dataset was split into train, validation, and test subsets. The validation and test sets remain fixed across all experiments in order to ensure fair comparisons between models.

To analyze the impact of labeled data availability, we constructed multiple training subsets from the golden training split, corresponding to four training regimes with 50, 100, 300, and 443 training images. Our method is specifically designed for low-data regime, so our primary experiments focus on the 50 and 100 training image settings. For scientific completeness only, we also include larger subsets (300 and 443 training images) to understand the behavior of our framework beyond its intended use case. The test and validation subsets remain fixed across all training regimes.

Importantly, this golden dataset remains untouched throughout the rest of the pipeline: no filtering, cleaning, or automated labeling procedures are applied to it. All preprocessing steps described in the following sections, such as label cleaning or automatic bounding box generation, are applied exclusively to the larger dataset collected from GIGA and other public sources.

\subsection{Automatic Filtering of Irrelevant Coordinates}

Unlike prior works that rely heavily on manual cleaning and annotation, our approach introduces an automated pipeline for both image filtering and bounding box generation.

The raw data set consists of school geographic coordinates obtained from GIGA sources. These coordinates (longitude, latitude) are intended to mark the location of a school, but they are limited to a single point per image, with no accompanying bounding boxes or segmentation masks. Beyond this lack of spatial annotation, the coordinates are sometimes noisy: they may be offset from the actual school building, occasionally falling on nearby roads, open fields, or other non-school areas. Through visual inspection of a large sample of school-centered images, we identified several common failure modes: some points were located over the ocean, forests, or deserts; others lacked any visible infrastructure or were affected by image quality issues such as blurriness or corruption.

To address these issues, we implemented an automatic cleaning pipeline based on three region-based metrics: vegetation ratio, desert ratio, and sea ratio. Each score is computed over a 200×200 pixel crop centered on the school location provided in the metadata. This localized crop enables us to assess the visual content most relevant to the target structure, while minimizing the influence of irrelevant background that may dominate the full 500×500 image. By restricting analysis to this focused region, we improve the robustness of the filtering process and reduce the likelihood of discarding valid school images located in complex environments, such as forests or coastal areas. Images that exceed any of the predefined thresholds for vegetation, desert, or sea coverage are excluded from training. More details are provided in Appendix~\ref{sec:AppendixB}.


In the following section, we describe how the cleaned image dataset is used to automatically generate bounding box annotations around school buildings through a segmentation-based pipeline.

\begin{figure}[t]
    \centering
    \begin{subfigure}[t]{0.32\linewidth}
        \includegraphics[width=\linewidth]{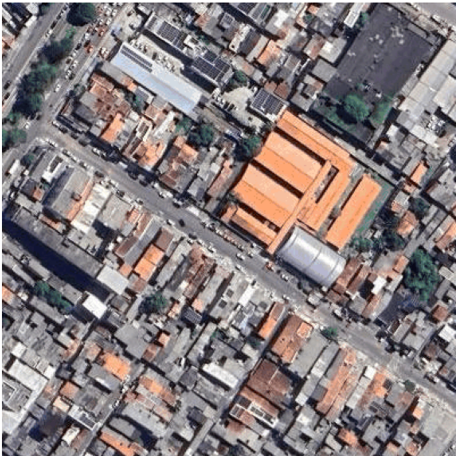}
        \caption{Initial image}
        \label{fig:crop400}
    \end{subfigure}
    \hfill
    \begin{subfigure}[t]{0.32\linewidth}
        \includegraphics[width=\linewidth]{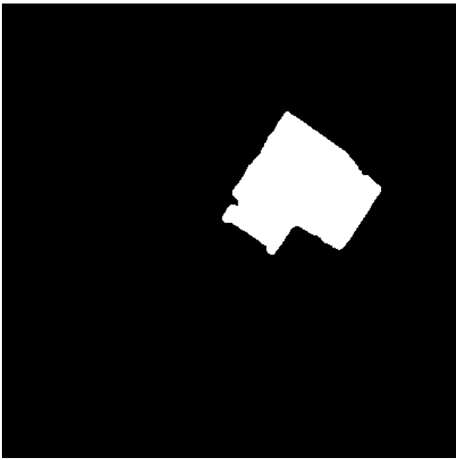}
        \caption{Mask}
        \label{fig:mask}
    \end{subfigure}
    \hfill
    \begin{subfigure}[t]{0.32\linewidth}
        \includegraphics[width=\linewidth]{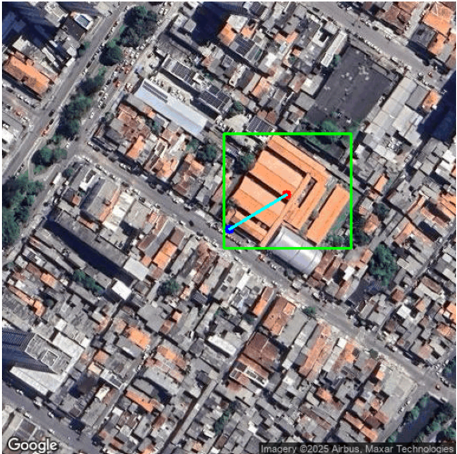}
        \caption{Generated BBox}
        \label{fig:bbox}
    \end{subfigure}
    \caption{Segmentation process: from crop to segmentation mask to generated bounding box.}
    \label{fig:segmentation_steps}
\end{figure}

\begin{table}[t]
\centering
\begin{tabular}{l c}
\toprule
\textbf{Category} & \textbf{U.S.} \\
\midrule

\textbf{Raw collected} & \\
\quad School-location images (no boxes)     & 12000 \\
\quad Non-school location images            & 2985 \\

\midrule
\textbf{Automatically labeled (with our pipeline)} & \\
\quad School images                         & 8703 \\
\quad Total auto-labeled (school + non-school) & 11688 \\

\bottomrule
\end{tabular}
\caption{\small Summary of raw and automatically labeled data for the U.S. dataset. These values are identical across all experimental settings.}
\label{table:usa_data_table}
\end{table}

\begin{table}[t]
\centering
\begin{tabular}{l cccc}
\toprule
\textbf{Category} & \textbf{50} & \textbf{100} & \textbf{300} & \textbf{443} \\
\midrule

\textbf{Golden Dataset} & & & & \\
\quad School images     & 32 & 65 & 195 & 288 \\
\quad Non-school images & 18 & 35 & 105 & 155 \\

\bottomrule
\end{tabular}
\caption{\small Composition of the Golden training dataset for the U.S. under different training regimes. Each column corresponds to a different number of manually labeled training samples.}
\label{table:usa_golden_variants}
\end{table}

\section{Methodology}

Following the automatic filtering of noisy and irrelevant aerial images, the next key challenge is generating reliable bounding box annotations. While manual labeling offers high accuracy, it is prohibitively time-consuming and labor-intensive particularly in the context of GIGA Project, where scalability is critical \citep{fu2021}.


Training object detection models solely on the small manually labeled dataset leads to suboptimal performance, both for simple detectors such as YOLOs and F.R-CNN and for models pretrained on large satellite and aerial imagery datasets like SatlasNet \citep{satlas}, as the number of annotated examples remains limited, as reported in Tables \ref{tab:all_results_tiny} and \ref{tab:results_golden_100}. This constraint is challenging in satellite and aerial imagery tasks, where the visual variability of buildings and environments requires large amounts of training data.

To address these limitations, we propose a scalable method that leverages weak supervision to generate school bounding box annotations from school center points with minimal human intervention. Our pipeline mainly consists of: (1) \textbf{Automatic Labeling}, which uses segmentation guided by the cleaned point annotations to generate bounding boxes; (2) \textbf{Model Training}, where object detectors are trained on the automatically labeled data; and (3) \textbf{Fine-Tuning}, where models are refined using a small, high-quality manually labeled subset to improve accuracy and correct systematic errors.

The intuition behind this pipeline is that automatically generated bounding boxes allow us to construct a large training dataset without extensive manual labeling. Because these annotations are produced automatically, they inevitably contain some level of noise. Nevertheless, this automatically labeled dataset provides sufficient supervision for an initial pretraining stage, enabling the model to learn a representation of school-related structures. In the final stage, this representation is refined through fine-tuning on a smaller, clean dataset containing high-quality manually annotated bounding boxes. As a result, the framework achieves strong detection performance while requiring only a limited amount of manual annotation.

Figure \ref{fig:diagram} shows an overview of the whole  pipeline.



\subsection{Automatic Labeling Pipeline} 

After automatic filtering of the aerial images, we generate bounding box annotations for school buildings using a segmentation-based approach. The process is guided by the school location points provided in the metadata, which approximate the center of each school within the image. These points serve as spatial priors, allowing the segmentation model to focus on the most relevant region while reducing computational overhead. Specifically, we extract a 400×400 pixel crop centered on each school location, which serves as input to the segmentation pipeline.
 

To segment potential school buildings within the cropped region, we leverage LangSAM \citep{langsam}, a variant of the Segment Anything Model (SAM) \citep{sam}. Built on top of SAM 2.0 \citep{ravi2024sam2} and GroundingDINO \citep{liu2024grounding}, LangSAM accepts both an image and a textual query, returning segmentation masks that correspond to the queried object class. This formulation is particularly well-suited to our setting, as it does not rely solely on precise location cues, but instead incorporates high-level semantic understanding of the image content to identify relevant structures.

In our pipeline, we provide LangSAM with the image crop and query it using three textual prompts: \textit{"building"}, \textit{"roof"} and \textit{"school"}. Using multiple prompts increases the robustness of the segmentation process. Indeed, relying on a single prompt may sometimes fail to produce a valid mask, By using several semantically related prompts, we increase the likelihood of detecting structures corresponding to school buildings or their roofs.

LangSAM returns a set of segmentation masks for each prompt, which we automatically filter to discard candidates that are either too small or excessively large, as these are unlikely to correspond to school building. Since the school location provided in the metadata approximates the center of the school in the image, we use this spatial prior to identify the most relevant candidate masks. Specifically, for each selected mask, we compute the Euclidean distance between its centroid and the school location at the center of the image. We retain the two closest masks and perform or not  a fusion step based on their spatial overlap.


After this, we validate the resulting mask using shape-based heuristics, including solidity, a metric known to be effective in identifying building-like structures in remote sensing imagery \citep{buiduing_detection}. Once a mask passes validation, we compute the bounding box that encloses its convex hull and use this as the final label for object detection training. If no valid mask is found, or if the centroid of the selected object is too far from the image center, the image is marked as invalid and rejected.

Despite requiring no human intervention, the proposed pipeline is capable of generating thousands of high-quality bounding box annotations, representing a substantial reduction in labeling cost compared to manual annotation. Table~\ref{table:usa_data_table} summarizes the number of bounding boxes generated through our segmentation-based labeling procedure.

While some degree of labeling noise is inevitable, the automatically generated labels cover a broad and diverse set of valid examples that would be prohibitively expensive to annotate manually. This highlights the practical scalability of our approach, particularly in data-scarce or resource-constrained regions. Figure~\ref{fig:segmentation_steps} illustrates an example of an a school aerial image alongside its segmentation mask produced by LangSAM and the final bounding box derived from that mask, demonstrating the pipeline’s ability to localize school-related structures in a weakly supervised setting.

In addition to the automatically generated bounding boxes for school images, we also incorporate non-school images without annotations. Together, these two components constitute our auto-labeled dataset: school images with automatically generated bounding boxes and background images containing no schools. This combination provides both positive and negative training examples while requiring no manual annotation. Final evaluation is performed on the golden test set, providing a robust measure of how well models trained on noisy, automatically generated labels generalize to high-quality ground truth.

\subsection{Model Training on Automatically Labeled Data}

Once the auto-labeled dataset is ready, we split  this data into training, validation, and test subsets. Particular attention is paid to data augmentation strategies. In contrast to \citep{tingzon2025}, which applied only rotations and flips, we incorporate translation as a key augmentation. This is critical for preventing the model from learning a degenerate solution in which predictions are biased toward the center of the image. By introducing translations, we ensure that the model learns to detect school-like features irrespective of their position, thereby improving generalization.

For model training, we evaluate several object detection architectures, including YOLO-based detectors \citep{yolov5_ultralytics, yolov8_ultralytics, yolov10, yolo11_ultralytics, yolo12_ultralytics}, Faster .R-CNN  (F.RCNN) \citep{fasterrcnn}, and the SatlasNet model pretrained on aerial imagery and high resolution imagery \citep{satlas}. YOLO and F.R-CNN models are initialized with pretrained weights COCO \citep{coco} and SatlasNet was pretrained on large scale aerial imagery   \citep{satlas}. We train these models on the automatically labeled dataset and  \textbf{\textit{evaluation is conducted exclusively on the manually labeled golden test set.}}
 To ensure unbiased evaluation, we enforce strict separation between the automatically labeled train set and the manually annotated validation and test sets, with no overlapping images between them.

\subsection{Fine-Tuning on Manually Labeled Data}

To further improve detection accuracy and bridge the gap between weak labels and high-quality ground truth, we fine-tune each detector using the manually labeled golden dataset. Our method is specifically designed for scenarios where manual annotations are scarce. Accordingly, our primary analysis focuses on low-data regimes (50–100 labeled images for training). We additionally report results using larger manually labeled training subsets (up to 443 images) to assess the behavior of the framework when more annotations are available

Fine-tuning is performed on each golden training subset, with the validation and test sets kept fixed across all experiments. This setup allows us to systematically evaluate our approach primarily under low-data conditions, while the large-data experiments serve as a reference.

This two-stage training strategy combines the scalability of weak supervision with the precision of manual annotations, enabling the models to correct systematic errors introduced during automatic labeling. As shown in the Results section, this approach consistently improves detection performance across different detector architectures and delivers strong gains precisely in the low-data regime.


We use ECP \citep{fourati2025every, fourati26ecpv2} for hyperparameter optimization during fine-tuning. Hyperparameter tuning in our setting is particularly challenging because each evaluation requires a full training run of the detection model, making the optimization process computationally expensive. As a result, the hyperparameter search problem can be viewed as a costly black-box optimization task, where the objective function can only be evaluated through time-consuming training and validation.

Given these constraints, exhaustive search strategies such as grid search are impractical, and even random search can quickly exceed the available computational budget. Therefore, an efficient search strategy is required to explore the hyperparameter space while minimizing the number of expensive evaluations. ECP provides a suitable solution in this context, as it is specifically designed to efficiently explore large and complex hyperparameter spaces under limited evaluation budgets. By guiding the search toward promising regions of the space, ECP reduces the need for costly exhaustive exploration and limits reliance on manual hyperparameter tuning.

More details about the ECP hyperparameter optimization procedure are provided in Appendix~\ref{sec:AppendixA}.

\color{black}

\section{Results}




\subsection{Experimental Setup}
 All training and fine-tuning experiments were performed using \textbf{ NVIDIA GTX 1080Ti GPU}. Fine-tuning on the golden dataset was complemented by automatic hyperparameter optimization using the ECP framework \citep{fourati2025every}, as described in Appendix~\ref{sec:AppendixA}. This ensured that each model configuration was tailored to its data characteristics without manual tuning.

We evaluate three detection models F.R‑CNN, SatlasNet, and YOLO26n under varying amounts of manually labeled data. YOLO26n was chosen from the YOLO family due to its superior performance compared to other YOLO models, as shown in Appendix \ref{sec:AppendixC}. Henceforth, we refer to YOLO26n simply as YOLO for brevity. For YOLO and F.R-CNN, we consider three training scenarios: (1) training only on the golden dataset, (2) training only on the automatically labeled dataset, and (3) our proposed approach (pretraining on auto-labeled data, then fine-tuning on golden data). SatlasNet, which is already pretrained on large-scale high-resolution aerial imagery, is trained only on the golden dataset (with and without ECP hyperparameter optimization) to assess its performance in the target domain. For evaluation, we use standard metrics from the computer vision literature: Precision, Recall, F1-score, mAP@50, and mAP@50:95. For Precision, recall and F1-score, a predicted bounding box is considered a true positive if its Intersection over Union (IoU) with a ground-truth box exceeds 0.5. More information about the metrics can be found in the Appendix \ref{sec:AppendixE}. All experiments are evaluated on the same fixed golden test set to ensure fair comparison across models and training scenarios. The results reported in Tables~\ref{tab:all_results_tiny} and~\ref{tab:results_golden_100} are the performance of each trained model on this fixed test set.

The key findings are as follows:
\begin{itemize}
\item Training only on the golden dataset yields limited performance due to data scarcity.
\item Good performance with detectors trained on auto-labeled data alone indicates effective representation learning.

\item In small-training-set regimes (50–100 training images), our two-stage strategy (pretrain on auto-labeled data, then fine-tune on golden data) substantially outperforms training on golden data alone.
\item \textbf{With only 50 golden images}, our YOLO variant achieves 0.705 mAP50 and 0.707 F1-score (Table~\ref{tab:all_results_tiny}) a gain of +0.403 mAP50 over training on golden data alone showing that we achieve high performance with our method using only 50 training images.
\item \textbf{With 100 golden images}, our YOLO26n model reaches 0.868 mAP50 and 0.817 F1-score (Table~\ref{tab:results_golden_100}) an improvement of +0.546 mAP50 over the golden-only baseline.
\end{itemize}

\subsection{Results on 50 Training Images}
We first evaluate the extreme case where only 50 manually labeled (golden) images are available for training. Validation and test sets contain 63 and 128 images, respectively, and are fixed across all experiments.

\textbf{Training only on golden data leads to poor performance due to data scarcity.} Training YOLO or F.R-CNN directly on these 50 golden images yields poor results especially in precision,  even with ECP hyperparameter tuning (Table~\ref{tab:all_results_tiny}), due to the small size of the training set. SatlasNet, despite being pretrained on large-scale aerial imagery, achieves high recall (0.833) but extremely low precision (0.019), resulting in an F1-score of only 0.037. ECP optimization improves SatlasNet's F1-score to 0.179 (Table~\ref{tab:results_satlas_tiny}), but this remains very limited. This behavior indicates that the detector produces a large number of false positives, as it detects all buildings not only schools because SatlasNet was originally pretrained to detect polygons (buildings) \citep{satlas} and continues to do so regardless of whether they are schools. Appendix \ref{sec:AppendixF} shows some examples of this behavior.

\textbf{Good performance with only auto-labeled data indicates that the model learns meaningful representations.} Training solely on the automatically labeled dataset (no golden images) and evaluating on the golden test set yields good performance: YOLO achieves 0.51 recall, and F.R-CNN reaches 0.60 mAP50 (Table~\ref{tab:all_results_tiny}). This surpasses SatlasNet’s precision and F1-score in this regime, despite SatlasNet seeing manual labels.  These findings highlight the strength of
the proposed automatic labeling strategy and indicate that the
detectors are able to learn meaningful representations from the
automatically generated annotations.

\textbf{Combining pretraining on auto-labeled data with fine-tuning on the golden dataset yields the best results.} As shown in Table~\ref{tab:all_results_tiny}, YOLO trained with our framework (auto pretrain + golden fine-tune) achieves 0.705 mAP50 and 0.707 F1-score. Compared to training only on the golden dataset, this is an average gain of approximately +0.38 across metrics. For F.R-CNN, the gain is even larger across all metrics (Table~\ref{tab:all_results_tiny}). An important observation is that our method yields the largest improvement on mAP@50:95, which emphasizes that the fine-tuning performed on the golden dataset significantly improves the quality and precision of the detections. These results demonstrate that weakly supervised pretraining enables strong detection performance using only 50 manually labeled samples, highlighting the label efficiency of the proposed framework and its suitability for scenarios where scalability is a major concern.

\begin{table}[t]
    \centering
\resizebox{\columnwidth}{!}{%
\begin{tabular}{lccccc}
\toprule
Model & mAP50 & Prec & Rec & F1 & mAP50:95 \\
\midrule
Yolo Golden +ECP & 0.302 & 0.289 & 0.440 & 0.349 & 0.144\\
YOLO Auto & 0.464 & 0.478 & 0.512 & 0.494& 0.144\\
YOLO Ours & \textbf{0.705} & \textbf{0.753} & \textbf{0.667} & \textbf{0.707}& \textbf{0.513} \\
\midrule
F.Rcnn Golden +ECP & 0.537& 0.242& 0.726& 0.363&0.194\\
F.Rcnn Auto & 0.600 & 0.193 & \textbf{0.869}  & 0.316 & 0.395\\
F.Rcnn Ours & \textbf{0.702} & \textbf{0.460} & 0.750& \textbf{0.570} & \textbf{0.429}\\

\bottomrule
\end{tabular}%
}
\caption{\small Comparison of object detection models in the 50-training-sample regime. \textit{YOLO Golden +ECP}  is trained directly on the Golden dataset using ECP hyperparameterization. \textit{YOLO Auto} is trained directly on the auto-labeled dataset. \textit{YOLO Ours}  corresponds to our proposed approach: the model is first trained on the auto-labeled dataset and then fine-tuned on the Golden training set. The same training strategies are applied to F.R-CNN models (\textit{Golden +ECP}, \textit{Auto}, and \textit{Ours} ). All models are evaluated on the same Golden test set. Results are reported in terms of mAP@50, Precision, Recall, F1-score, and mAP@50:95.}
\label{tab:all_results_tiny}
\end{table}

\begin{table}[t]
    \centering
\resizebox{\columnwidth}{!}{%
\begin{tabular}{lccccc}
\toprule
Model & mAP50 & Prec & Rec & F1 & mAP50:95 \\
\midrule
SatlasNet (no ECP) & 0.583 & 0.019 & 0.833 & 0.037& 0.213 \\
SatlasNet +ECP & 0.493 & 0.103 & 0.690 & 0.179 &0.201\\
\bottomrule
\end{tabular}%
}
\caption{\small Comparison of SatlasNet in the 50-training-sample regime. \textit{SatlasNet (no ECP)} was trained directly on the golden data without ECP hyperparametrization, and \textit{SatlasNet +ECP} was trained with ECP. The two models are evaluated on the fixed golden test set. Results are reported in terms of mAP50, precision, recall, F1-score, and mAP50:95.}
\label{tab:results_satlas_tiny}
\end{table}

\subsection{Results on 100 Training Images}
We next increase the golden training set to 100 images, keeping validation and test sets unchanged.

\textbf{Golden-only training remains limited.} YOLO and F.R-CNN still perform poorly when trained on this regime (Table~\ref{tab:results_golden_100}). SatlasNet still achieves high recall but very low precision (0.035), yielding an F1-score of only 0.067 (Table~\ref{tab:results_satlas_golden_100}). This confirms the previous observation: SatlasNet's pretraining continues to bias its predictions toward detecting all buildings, regardless of whether they are schools. 

\textbf{ECP improves SatlasNet but precision remains an issue.} Hyperparameter optimization using ECP with F1-score as objective raises SatlasNet’s precision to 0.552 and F1-score to 0.679 (Table~\ref{tab:results_satlas_golden_100}). However, the precision-recall balance is still suboptimal.

\textbf{Our two-stage strategy again yields substantial improvements.} As reported in Table~\ref{tab:results_golden_100}, YOLO with our approach achieves 0.868 mAP50 and 0.817 F1-score an average gain of approximately +0.5 across all metrics compared to golden-only training. mAP50 improves by more than +0.54. F.R-CNN shows similar gains in all metrics especially for  F1-score and  mAP50. These results confirm that our framework is highly label-efficient, achieving strong performance with only 100 manual annotations.

\begin{figure}[t]
\centering
\includegraphics[width=\columnwidth]{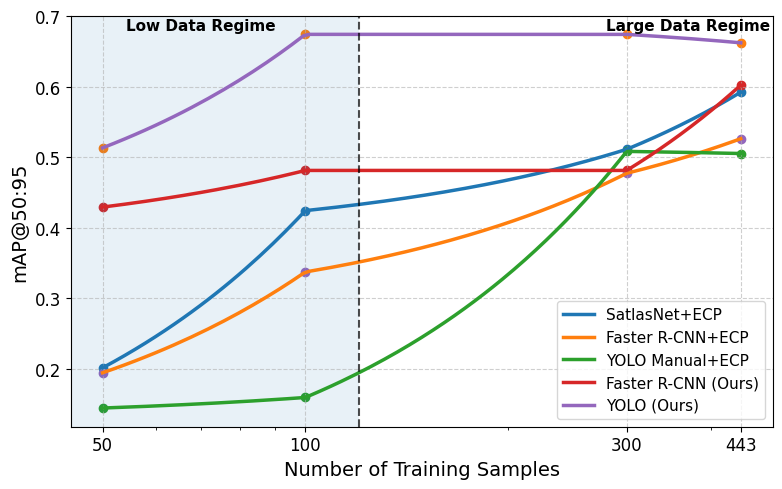}
\caption{\small mAP@50:95 performance of different detectors and training strategies across different training regimes (50, 100, 300, and 443 samples).}
\label{fig:training_regime_map5095}
\end{figure}

\begin{figure}[t]
\centering
\includegraphics[width=\columnwidth]{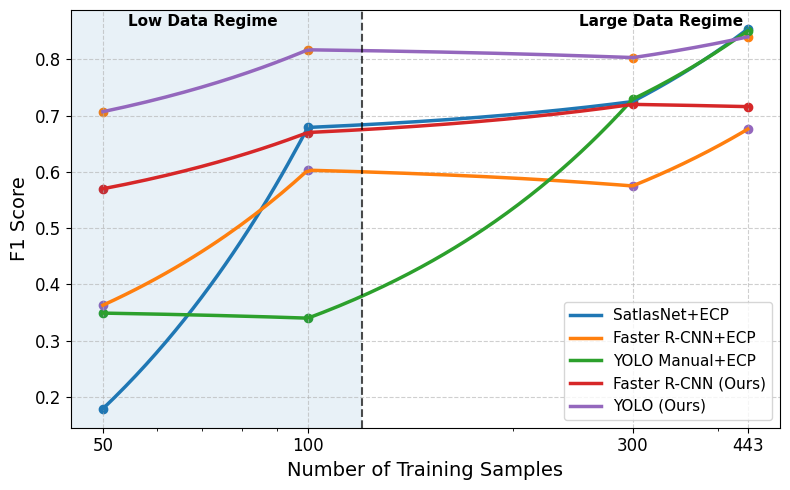}
\caption{\small F1-score@50 performance of different detectors and training strategies across different training regimes (50, 100, 300, and 443 samples).}
\label{fig:training_regime_F1}
\end{figure}

It is also important to note that our approach yields a gain of 0.1 in YOLO compared to the previous regime, which is expected. However, it should be observed that the results obtained with our approach in this regime are highly competitive with the best results reported in the large‑data regime, as shown in Figures \ref{fig:training_regime_F1} and \ref{fig:training_regime_map5095}. With only 100 training images, we achieve competitive performance relative to large‑data regimes that require substantially more labeled data.

Beyond the quantitative gains, the results of this regime provide important insight into the role of weakly supervised pretraining. The automatically generated annotations appear sufficient to guide the model toward learning meaningful visual features associated with school structures, while the small golden dataset primarily serves to refine these representations and correct systematic localization errors. This combination leads to substantial performance gains in low-data regimes and demonstrates that a large portion of the representation learning can be effectively achieved using weak supervision. As a result, the proposed approach significantly reduces the dependence on extensive manually annotated datasets, enabling a scalable solution for large-scale mapping tasks where annotation resources are limited



\begin{table}[t]
    \centering
\resizebox{\columnwidth}{!}{%
\begin{tabular}{lccccc}
\toprule
Model & mAP50 & Prec & Rec & F1 & mAP50:95\\
\midrule
Yolo Golden + ECP & 0.322 & 0.281 &0.429 &0.340 & 0.159\\
YOLO Auto & 0.464 & 0.478 & 0.512 & 0.494& 0.332 \\
YOLO Ours & \textbf{0.868} & \textbf{0.782} & \textbf{0.855} & \textbf{0.817}& \textbf{0.674} \\
\midrule
F.Rcnn Golden + ECP & 0.587& 0.522 & 0.714 &0.603 & 0.337\\
F.Rcnn Auto & 0.600 & 0.193 & \textbf{0.869} & 0.316& 0.395\\
F.Rcnn Ours & \textbf{ 0.712} & \textbf{0.631} & 0.714 & \textbf{0.670} &  \textbf{0.481}\\
\bottomrule
\end{tabular}%
}
\caption{\small Comparison of object detection models in the 100-training-sample regime. \textit{YOLO Golden +ECP}  is trained directly on the Golden dataset using ECP hyperparameterization. \textit{YOLO Auto} is trained directly on the auto-labeled dataset. \textit{YOLO Ours}  corresponds to our proposed approach: the model is first trained on the auto-labeled dataset and then fine-tuned on the Golden training set. The same training strategies are applied to F.R-CNN models (\textit{Golden +ECP}, \textit{Auto}, and \textit{Ours} ). All models are evaluated on the same Golden test set. Results are reported in terms of mAP@50, Precision, Recall, F1-score, and mAP@50:95.}
    \label{tab:results_golden_100}
\end{table}

\begin{table}[t]
    \centering
\resizebox{\columnwidth}{!}{%
\begin{tabular}{lccccc}
\toprule
Model & mAP50 & Prec & Rec & F1 & mAP50:95\\
\midrule
SatlasNet (no ECP) & 0.717 & 0.035 & \textbf{0.893}  & 0.067 & 0.348\\
SatlasNet +ECP & \textbf{0.823} & \textbf{0.552}& 0.881 & \textbf{0.679}& \textbf{0.424}\\
\bottomrule
\end{tabular}%
}
\caption{\small Comparison of SatlasNet in the 100-training-sample regime. \textit{SatlasNet (no ECP)} was trained directly on the golden data without ECP hyperparametrization, and \textit{SatlasNet + ECP} was trained with ECP. The two models are evaluated on the fixed golden test set. Results are reported in terms of mAP50, precision, recall, F1-score, and mAP50:95.}
    \label{tab:results_satlas_golden_100}
\end{table}

\subsection{Additional Findings}

Our method is designed for low-data regime. Experiments with larger manually labeled datasets are included only for scientific completeness and do not represent the intended use case of our framework.

\textbf{Results on 300 training samples (Appendix~\ref{sec:AppendixD}).} Our two-stage strategy continues to improve performance for both YOLO26n and F.R-CNN compared to training only on the golden dataset. Gains remain significant, particularly in mAP@50:95 and precision. However, the improvements are less pronounced than in the 50 and 100 sample settings. For example, the average gain over golden-only training for YOLO26n drops from approximately 0.30 (at 50 samples) to around 0.15 (at 300 samples). This confirms that pretraining on auto-labeled data provides meaningful feature representations, but the marginal benefit diminishes as more manual labels become available.

In this regime, SatlasNet trained only on the golden dataset achieves competitive results across most metrics. YOLO26n and F.R-CNN also perform well when trained solely on manual data. However, it is important to note that mAP@50:95 is still significantly improved when using our approach with YOLO as reported in Table \ref{tab:all_results_300} and Figure \ref{fig:training_regime_map5095}, highlighting its effectiveness in enhancing the overall precision and quality of the detections.

\textbf{Results on 443 training samples (Appendix \ref{sec:AppendixD}).} When the golden training set is increased to 443 images, the advantage of our approach becomes minimal. SatlasNet trained only on the golden dataset reaches very strong performance, also YOLO26n trained only on golden achieves scores exceeding 0.8 across all evaluation metrics. Our two-stage strategy still performs well but does not provide significant gains over training on manual data alone, except for the map50:95 where our approach achieves the best performance (Figure \ref{fig:training_regime_map5095}).

\textbf{Limitation and takeaway.} Our method is specifically designed for low-resource settings, where manual annotation is the primary bottleneck. In this regime (typically 50–100 labeled images) it delivers its strongest and most practical benefits. As expected, increasing the amount of labeled data improves any detector, including our detectors. However, even if that scenario is not the focus of our work, we observe consistent improvements in mAP@50:95, indicating better localization quality. Off-the-shelf pretrained models like SatlasNet can become competitive when hundreds or thousands of labels are available, but that setting contradicts the scalability constraints of initiatives like GIGA. Thus, our core contribution remains label efficiency precisely where it matters most: when manual labels are scarce. With only 50–100 manually annotated samples, our approach achieves performance comparable to models trained on 443 samples, making it highly relevant for large-scale mapping applications.

\section{Conclusion and Future Work}

In this work, we introduced a weakly supervised framework for object detection in aerial and satellite imagery, specifically designed to address the constraints of limited manual annotations. By combining large-scale automatically labeled data with a small high-quality manually labeled dataset through a two-stage training strategy, we demonstrate that high detection performance can be achieved even in extremely low-data regimes.

Our experiments show that the proposed approach consistently improves performance across different detector architectures, including YOLO and F.R-CNN. Notably, in the most constrained setting with only 50 manually labeled images, our method leads to substantial gains across all metrics, highlighting its robustness and general applicability. This demonstrates that the benefits of our framework are not model-specific, but rather stem from the quality of the learned representations during the pretraining phase. The representations learned during this phase enable the model to generalize effectively, and the subsequent fine-tuning on a small golden dataset allows correcting systematic errors introduced by noisy labels. This combination proves to be a highly efficient strategy for balancing scalability and accuracy.

Our analysis also confirms that this label-efficient design is particularly valuable compared to off-the-shelf pretrained detectors such as SatlasNet, which underperform in low-data settings due to insufficient fine-tuning data. While such models can achieve strong results when more labels become available, that scenario is not the focus of our work. In the context of global scalability such as the GIGA initiative, where manual annotation must be minimized our approach offers a more practical and immediately effective solution.

Overall, this work demonstrates that accurate and scalable object detection can be achieved with minimal manual supervision. Requiring as few as 50 to 100 labeled images, our method drastically reduces annotation costs while maintaining strong performance, making it particularly suitable for large-scale deployments in real-world settings.

Future work will focus on extending this framework to additional countries and diverse geographic contexts, as well as improving the robustness of the automatic labeling pipeline to further reduce noise and enhance generalization. Beyond detection, we plan to leverage the resulting school location data to support school connectivity analysis, enabling the identification of connectivity gaps and helping inform data-driven planning of digital infrastructure. Such information can play a key role in initiatives aimed at connecting schools to reliable internet access and supporting equitable digital education


\bibliography{aaai2026}

\appendices
\section{Automatic Hyperparameter Tuning using ECP}
\label{sec:AppendixA}

In this appendix, we describe the hyperparameter optimization strategy applied to models fine-tuned on the manually labeled dataset (golden data). 

For hyperparameter optimization, we employed the \textbf{Every Call is Precious (ECP)} algorithm \citep{fourati2025every, fourati26ecpv2}, a global optimization method tailored for expensive, black-box functions, and computationally expensive objective functions. 
ECP has been shown to outperform prior state-of-the-art methods under tight evaluation budgets, making it particularly practical when dealing with expensive functions.

In our experiments, each function call corresponds to a full fine-tuning run of the detector using a specific hyperparameter configuration. Each run included up to a 50 times of fine-tuning steps, with early stopping enabled to reduce unnecessary computation.

ECP was used to tune some hyperparameters depending on the model. For each model we provide  for a full list of tuned hyperparameters and their respective search ranges per model is provided in Tables \ref{table:ecp_bounds_yolo} and \ref{tab:ecp_bounds_rcnn} and \ref{tab:ecp_bounds_satlas}.
Despite the limited budget, ECP consistently discovered configurations that led to good improvements in all metrics compared to default settings.
These findings highlight the critical role of hyperparameter optimization in improving detection performance.

\begin{table}[ht]
\centering

\begin{tabular}{llc}
\toprule
\textbf{Index} & \textbf{Hyperparameter} & \textbf{Search Range} \\
\midrule
1 & Initial learning rate ($lr0$) & $[10^{-4}, 10^{-2}]$ \\
2 & Learning rate factor ($lrf$) & $[0.01,\ 0.1]$ \\
3 & Momentum & $[0.90,\ 0.98]$ \\
4 & Weight decay & $[10^{-5},\ 0.005]$ \\
5 & Box loss gain & $[7.0,\ 10.0]$ \\
6 & Translate augmentation & $[0.0,\ 0.3]$ \\
7 & Classification loss gain ($cls$) & $[0.2,\ 1.5]$ \\
8 & DFL loss gain & $[0.8,\ 2.5]$ \\
9 & Dropout rate & $[0.0,\ 0.4]$ \\
10 & Random erasing probability & $[0.1,\ 0.5]$ \\
\bottomrule
\end{tabular}
\caption{Hyperparameters tuned with ECP and their corresponding search intervals.}
\label{table:ecp_bounds_yolo}
\end{table}


\begin{table}[ht]
\centering

\begin{tabular}{llc}
\toprule
\textbf{Index} & \textbf{Hyperparameter} & \textbf{Search Range} \\
\midrule
1 & Learning rate & $[10^{-5}, 5\times10^{-3}]$ \\
2 & Momentum & $[0.85,\ 0.98]$ \\
3 & Weight decay & $[10^{-6},\ 10^{-3}]$ \\
4 & RPN NMS threshold  & $[0.4,\ 0.8]$ \\
5 & Box score threshold  & $[0.2,\ 0.5]$ \\
6 & Box NMS threshold  & $[0.4,\ 0.7]$ \\
\bottomrule
\end{tabular}

\caption{Hyperparameters tuned with ECP for the F.R-CNN model and their corresponding search intervals.}
\label{tab:ecp_bounds_rcnn}

\end{table}



\begin{table}[ht]
\centering

\begin{tabular}{llc}
\toprule
\textbf{Index} & \textbf{Hyperparameter} & \textbf{Search Range} \\
\midrule
1 & Learning rate & $[10^{-5}, 5\times10^{-4}]$ \\
2 & Weight decay & $[10^{-6}, 10^{-3}]$ \\
3 & Confidence threshold  & $[0.01,\ 0.4]$ \\
4 & NMS threshold & $[0.4,\ 0.8]$ \\
\bottomrule
\end{tabular}

\caption{Hyperparameters tuned with ECP for the SatlasNet model and their corresponding search intervals.}
\label{tab:ecp_bounds_satlas}
\end{table}


\section{Boxplots Distributions of Outliers: All metrics}
\label{sec:AppendixB}
This appendix presents details for three categories of automatically identified outliers: \textbf{vegetation}, \textbf{sea}, and \textbf{desert}, across U.S.

To determine appropriate thresholds for each filtering metric, we sampled 1,000 randomly selected images and computed vegetation, desert, and sea ratios. The thresholds were then calibrated using the 95th percentile of the corresponding distribution, allowing the filtering process to account for regional environmental variation.

\begin{figure}[t]
    \centering
    \begin{subfigure}[t]{0.32\linewidth}
        \includegraphics[width=\linewidth]{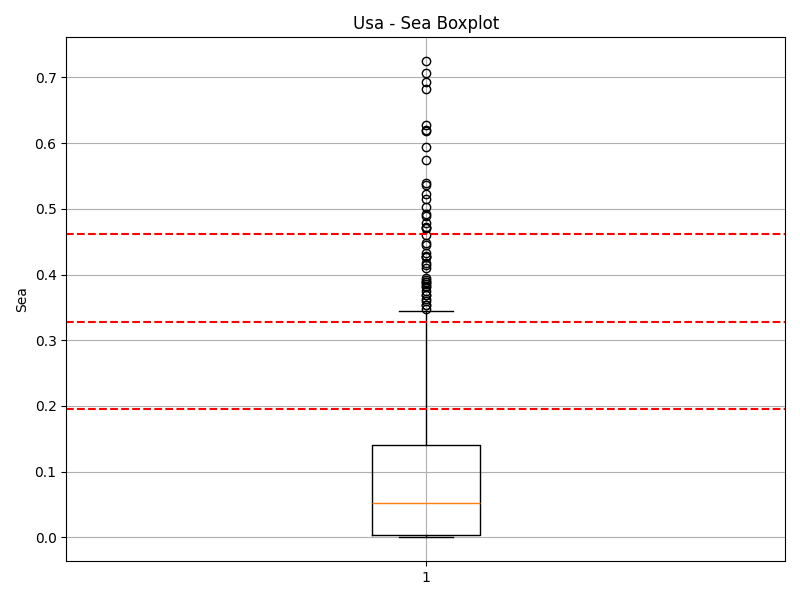}
        \caption{sea boxplot}
        \label{fig:sea}
    \end{subfigure}
    \hfill
    \begin{subfigure}[t]{0.32\linewidth}
        \includegraphics[width=\linewidth]{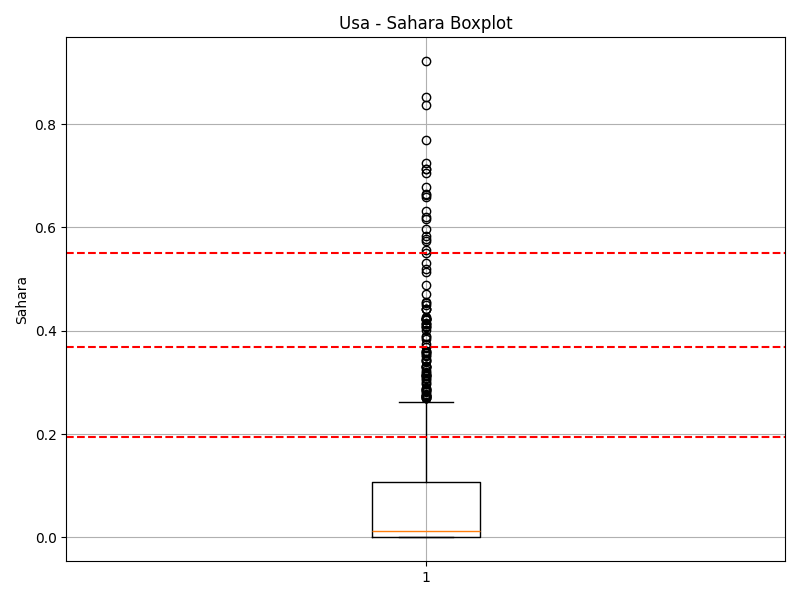}
        \caption{desert boxplot}
        \label{fig:desert}
    \end{subfigure}
    \hfill
    \begin{subfigure}[t]{0.32\linewidth}
        \includegraphics[width=\linewidth]{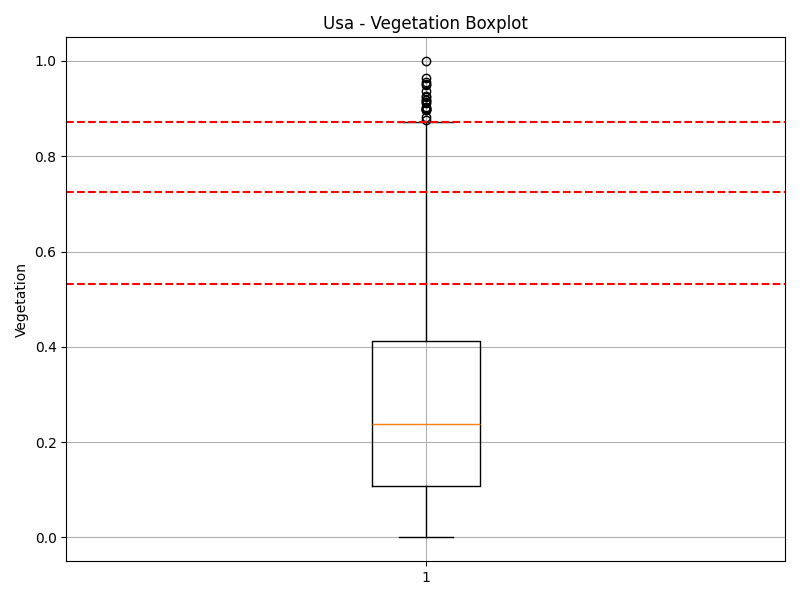}
        \caption{vegetation boxplot}
        \label{fig:vegetation}
    \end{subfigure}
    \caption{U.S. Boxplots}
    \label{fig:boxplots}
\end{figure}

The boxplots in Figures \ref{fig:sea}, \ref{fig:desert} and \ref{fig:vegetation} provide insights about the distribution of the three metrics used. They also support the choice of thresholds used during the data cleaning process.

For each metric, thresholds were computed independently . Specifically, we used the 95\textsuperscript{th} percentile of the score distribution as a data-driven estimate of extreme values, and applied a conservative threshold defined as:

\[
\text{Threshold} = \max(0.8, \text{percentile}_{95})
\]

This strategy ensures that only the most extreme outliers are removed, while preserving relevant images that may exhibit moderate levels of vegetation, sea, or desert features. The fixed lower bound of 0.8 was chosen based on empirical calibration.

\section{Benchmarking YOLOs}
\label{sec:AppendixC}

To select the most suitable YOLO variant for our experiments, we benchmarked several YOLO architectures (v5n, v8n, v10n, v11n, v12n, v26n) by training each on the golden dataset and evaluating on the golden test set. For the low-data regime (100 training samples), YOLOv26n consistently achieved the best performance across most metrics, including F1-score (0.491) and mAP50:95 (0.278), as shown in Table~\ref{tab:yolo_detailed_comparison_smaller}. Therefore, YOLOv26n was selected for all low-data experiments. For the larger dataset (300 training samples), YOLOv26n again showed strong results (Table~\ref{tab:yolo_detailed_comparison_small}), reaching the highest mAP@50 (0.659) and F1-score (0.519), confirming its robustness across data regimes. 

\begin{table}[H]
\centering
\begin{tabular}{l|ccccc}
\toprule
\textbf{Model} & \textbf{Prec} & \textbf{Rec} & \textbf{F1} & \textbf{mAP50} & \textbf{mAP50:95} \\
\midrule
YOLOv5n   & 0.367 & 0.560 & 0.443 & \textbf{0.528} & 0.235 \\
YOLOv8n   & 0.281 & 0.429 & 0.340 & 0.298 & 0.124 \\
YOLOv10n  & 0.398 & 0.607 & 0.481 & 0.452 & 0.217 \\
YOLOv11n  & 0.273 & 0.417 & 0.330 & 0.299 & 0.165 \\
YOLOv12n  & 0.383 & 0.583& 0.462 & 0.440 & 0.196 \\
YOLOv26n  & \textbf{0.406} &\textbf{0.619} & \textbf{0.491} & 0.490 & \textbf{0.278} \\

\bottomrule
\end{tabular}
\caption{\small Detailed comparison of YOLO variants on the U.S. 100 training golden dataset. Best result for each metric is in bold.}
\label{tab:yolo_detailed_comparison_smaller}
\end{table}

\begin{table}[H]
\centering
\resizebox{\columnwidth}{!}{%
\begin{tabular}{l|ccccc}
\toprule
\textbf{Model} & \textbf{Prec} & \textbf{Rec} & \textbf{F1} & \textbf{mAP@50} & \textbf{mAP@50:95} \\
\midrule
YOLOv5n   & 0.414 & 0.631 & 0.500 & 0.507 & 0.235 \\

YOLOv8n   & 0.367 & 0.560 & 0.443 & 0.476 & 0.227 \\

YOLOv10n  & 0.203 & 0.310 & 0.245 & 0.221 & 0.061 \\

YOLOv11n  & \textbf{0.471} & 0.393 & 0.428 & 0.368 & 0.153 \\

YOLOv12n  & 0.344 & 0.524 & 0.415 & 0.416 & 0.161 \\

YOLOv26n  & 0.430 &\textbf{0.655} & \textbf{0.519} & \textbf{0.659} & \textbf{0.254 }\\

\bottomrule
\end{tabular}%
}
\caption{\small Detailed comparison of YOLO variants on the U.S. 300 training golden dataset. Best result for each metric is in bold.}
\label{tab:yolo_detailed_comparison_small}
\end{table}

\section{Results on large data regime}
\label{sec:AppendixD}
 In the large data regime, we use the same detection models as in the low-data regime: F.R-CNN, SatlasNet, and YOLO. For YOLO, we conducted a preliminary benchmarking across multiple variants of YOLO to identify the most suitable model for this data regime.

The detailed Yolo benchmarking results and model comparison are provided in Appendix C. 

We then evaluate the proposed approach under two large-data scenarios:  a large data setting with 300 training images, and a big-data setting with 442 training images. The validation and test sets are the same as in the low-data regime.

\subsubsection{Results on small golden (300 train images)}

In this experimental scenario, the size of the golden training dataset to 300 images while keeping the validation and test sets unchanged. This setup allows us to analyze how the different detectors scale when a moderate amount of labeled data becomes available.



\small
\begin{table}[t]
\centering
\resizebox{\columnwidth}{!}{%
\begin{tabular}{lccccc}
\toprule
Model & mAP50 & Prec & Rec & F1& mAP50:95 \\
\midrule
SatlasNet & \textbf{0.891} & 0.590 & \textbf{0.941}  & 0.725& 0.511 \\
\midrule
YOLO Golden +ECP& 0.802 & 0.714 & \textbf{0.744} &0.729 &0.508 \\
YOLO Auto & 0.464 & 0.478 & 0.512 & 0.494& 0.332 \\
YOLO Ours & \textbf{0.863} & \textbf{0.874} & 0.742 & \textbf{0.803}& \textbf{0.674} \\
\midrule
F.Rcnn Golden +ECP& 0.712 &0.442& 0.821&0.575& 0.477\\
F.Rcnn Auto  & 0.600 & 0.193 & \textbf{0.869} & 0.316& 0.395\\
F.Rcnn Ours & \textbf{0.775} & \textbf{0.657} & 0.798 & \textbf{0.720} & \textbf{0.481} \\

\bottomrule
\end{tabular}%
}
\caption{\small Comparison of object detection models in the 300-training-sample regime. SatlasNet is trained directly on the Golden dataset. \textit{YOLO Golden +ECP}  is trained directly on the Golden dataset using ECP hyperparameterization. \textit{YOLO Auto} is trained directly on the auto-labeled dataset. \textit{YOLO Ours}  corresponds to our proposed approach: the model is first trained on the auto-labeled dataset and then fine-tuned on the Golden training set. The same training strategies are applied to F.R-CNN models (\textit{Golden +ECP}, \textit{Auto}, and \textit{Ours} ). All models are evaluated on the same Golden test set. Results are reported in terms of mAP@50, Precision, Recall, F1-score, and mAP@50:95.}
\label{tab:all_results_300}
\end{table}

\subsubsection{Results on 443 train images}

In the final experimental setting, we train the detectors on the full Golden dataset, which contains 443 manually labeled training images. The validation and test sets remain unchanged in order to preserve the comparability of the experiments.

\begin{table}[t]
\centering
\resizebox{\columnwidth}{!}{%
\begin{tabular}{lccccc}
\toprule
Model & mAP50 & Prec & Rec & F1 & mAP50:95 \\
\midrule
SatlasNet & 0.876 & 0.809 & \textbf{0.905} & \textbf{0.854}& 0.592 \\
\midrule
YOLO Golden +ECP & \textbf{0.891} & 0.884 & \textbf{0.819} & \textbf{0.851}& 0.505\\
YOLO Auto & 0.464 & 0.478 & 0.512 & 0.494&0.332 \\
YOLO Ours & 0.876 & \textbf{0.893}  & 0.793 & 0.840& \textbf{0.662} \\
\midrule
F.Rcnn Golden +ECP& 0.786 &  0.569&0.833& 0.676& 0.526\\
F.Rcnn Auto  & 0.600 & 0.193 & \textbf{0.869} & 0.316 & 0.395 \\
F.Rcnn Ours & \textbf{0.780} & \textbf{0.642} & 0.810 & \textbf{0.716} & \textbf{0.602}\\
\bottomrule
\end{tabular}%
}
\caption{\small Comparison of object detection models in the 443-training-sample regime. SatlasNet is trained directly on the Golden dataset. \textit{YOLO Golden +ECP}  is trained directly on the Golden dataset using ECP hyperparameterization. \textit{YOLO Auto} is trained directly on the auto-labeled dataset. \textit{YOLO Ours}  corresponds to our proposed approach: the model is first trained on the auto-labeled dataset and then fine-tuned on the Golden training set. The same training strategies are applied to F.R-CNN models (\textit{Golden +ECP}, \textit{Auto}, and \textit{Ours} ). All models are evaluated on the same Golden test set. Results are reported in terms of mAP@50, Precision, Recall, F1-score, and mAP@50:95.}
\label{tab:results_golden_442}
\end{table}

\section{Metrics}
\label{sec:AppendixE}
We use the following standard metrics to evaluate object detection performance. Their formal definitions are provided below.

\paragraph{Precision}
Precision measures the accuracy of the model’s predictions by quantifying the proportion of correctly predicted positive samples among all predicted positives. It is defined as:

\begin{equation}
\text{Precision} = \frac{TP}{TP + FP}
\end{equation}

where $TP$ denotes the number of true positives and $FP$ the number of false positives.

\paragraph{Recall}
Recall evaluates the ability of the model to detect all relevant objects. It is defined as the proportion of actual positive samples that are correctly identified:

\begin{equation}
\text{Recall} = \frac{TP}{TP + FN}
\end{equation}

where $FN$ denotes the number of false negatives.

\paragraph{F1-score}
The F1-score provides a balance between Precision and Recall and is defined as the harmonic mean of the two:

\begin{equation}
F1 = 2 \cdot \frac{\text{Precision} \cdot \text{Recall}}{\text{Precision} + \text{Recall}}
\end{equation}

\paragraph{Mean Average Precision (mAP)}
The mean Average Precision (mAP) summarizes the overall detection performance across all classes. It is computed as the mean of the Average Precision (AP) over all classes:

\begin{equation}
mAP = \frac{1}{C} \sum_{c=1}^{C} AP(c)
\end{equation}

where $C$ is the number of classes and 
As defined in \citep{everingham2010pascal}, Average Precision $AP(c)$ measures the area under the precision-recall curve for class c, and mean Average Precision (mAP) is its average across classes. mAP@50:95 extends this by computing AP at multiple IoU thresholds (from 0.50 to 0.95 in increments of 0.05), providing a stricter evaluation of spatial alignment between predicted and ground-truth.

\section{Failure Cases of SatlasNet}
\label{sec:AppendixF}

In this appendix, we present qualitative examples illustrating failure cases of SatlasNet when trained with only 50 labeled training images. As discussed in the main paper, SatlasNet is pretrained on large-scale satellite imagery with annotations primarily focused on polygonal object detection. While this pretraining provides strong general visual representations, it can also introduce biases when applied to tasks with different target objects.

In the case of school detection, we observe that the model frequently produces false positives on structures that exhibit geometric patterns similar to the polygonal objects seen during pretraining. In particular, buildings or areas with regular shapes or rectangular layouts are sometimes incorrectly detected as schools.

\begin{figure}[H]
\centering
\begin{subfigure}{0.48\columnwidth}
    \centering
    \includegraphics[width=\linewidth]{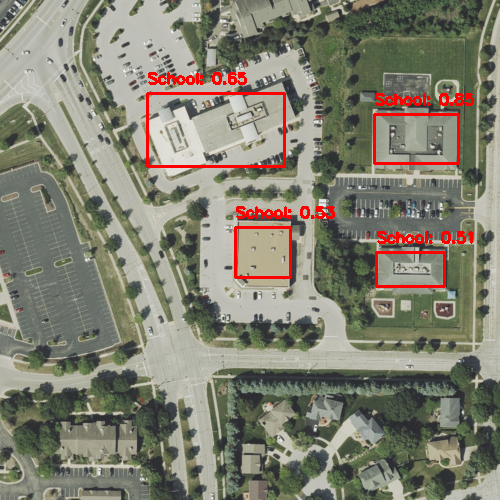}
    \caption{}
    \label{fig:satlas_failure1}
\end{subfigure}
\hfill
\begin{subfigure}{0.48\columnwidth}
    \centering
    \includegraphics[width=\linewidth]{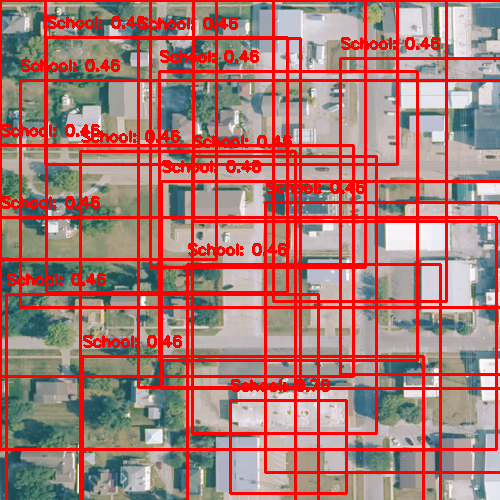}
    \caption{}
    \label{fig:satlas_failure2}
\end{subfigure}

\caption{\small Examples of false positives produced by SatlasNet trained with 50 images. 
The model incorrectly detects schools due to geometric structures that resemble polygonal objects observed during SatlasNet pretraining. Such errors highlight the bias induced by polygon-based pretraining in low-data regimes.}
\label{fig:satlas_failures}
\end{figure}

Figures~\ref{fig:satlas_failure1} and \ref{fig:satlas_failure2} illustrate two representative examples of such failure cases. In both images, SatlasNet incorrectly predicts the presence of a school in regions that correspond to other types of structures. 

\end{document}